\documentclass[runningheads]{llncs}
\usepackage[T1]{fontenc}
\usepackage{graphicx}
\usepackage{booktabs}
\usepackage[misc]{ifsym}
\newcommand{\corr}{(\Letter)}
\usepackage{mwe}

\usepackage{subcaption}
\usepackage{tabularx}
\usepackage{booktabs}
\usepackage{xurl}
\usepackage{algorithm}
\usepackage{algorithmic}
\usepackage{amsmath}
\usepackage{booktabs}
\usepackage{enumerate}
\usepackage{multirow}
\usepackage{upgreek}
\usepackage{mathtools}
\usepackage{amsfonts}

\graphicspath{{fig/}}

\begin{document}

\title{Long-term Fairness in Ride-Hailing Platform}

\titlerunning{Long-term Fairness in Ride-Hailing Platform}


\author{Yufan Kang\inst{1} \and
Jeffrey Chan\inst{1} \corr \and
Wei Shao\inst{2} \and Flora D. Salim\inst{3} \and Christopher Leckie\inst{4}}

\authorrunning{Published as a Conference Paper in ECML PKDD}

\tocauthor{Yufan Kang, Jeffrey Chan, Wei Shao, Flora D. Salim, Christopher Leckie}
\toctitle{Long-term Fairness in Ride-Hailing Platform}

\institute{RMIT University, Melbourne, Australia \email{yufan.kang@student.rmit.edu.au, jeffrey.chan@rmit.edu.au}
\and
Data61, CSIRO, Clayton, Australia \email{phdweishao@gmail.com}
\and
University of New South Wales, Sydney, Australia
\email{flora.salim@unsw.edu.au}
\and
The University of Melbourne, Carlton, Australia
\email{caleckie@unimelb.edu.au}
}

\maketitle              

\begin{abstract}
Matching in two-sided markets such as ride-hailing has recently received significant attention. However, existing studies on ride-hailing mainly focus on optimising efficiency, and fairness issues in ride-hailing have been neglected. Fairness issues in ride-hailing, including significant earning differences between drivers and variance of passenger waiting times among different locations, have potential impacts on economic and ethical aspects.  The recent studies that focus on fairness in ride-hailing exploit traditional optimisation methods and the Markov Decision Process to balance efficiency and fairness. However, there are several issues in these existing studies, such as myopic short-term decision-making from traditional optimisation and instability of fairness in a comparably longer horizon from both traditional optimisation and Markov Decision Process-based methods. To address these issues, we propose a dynamic Markov Decision Process model to alleviate fairness issues currently faced by ride-hailing, and seek a balance between efficiency and fairness, with two distinct characteristics: (i) a prediction module to predict the number of requests that will be raised in the future from different locations to allow the proposed method to consider long-term fairness based on the whole timeline instead of consider fairness only based on historical and current data patterns; (ii) a customised scalarisation function for multi-objective multi-agent Q Learning that aims to balance efficiency and fairness. Extensive experiments on a publicly available real-world dataset demonstrate that our proposed method outperforms existing state-of-the-art methods.  

\keywords{Applications\and Fairness \and Human-aware Planning and Scheduling.}
\end{abstract}

\section{Introduction}
Ride-hailing systems have become increasingly prevalent as a mode of transportation, with platforms, such as Uber, utilising artificial intelligence (AI) algorithms to match drivers to passengers efficiently \cite{mohlmann2019people}. However, while these algorithms succeed in optimising earnings for drivers and reducing passenger waiting time, they often result in inequities, such as wage disparities based on gender or race \cite{cook2021gender}. As a result, there is a growing interest in the fair allocation of jobs to drivers in ride-hailing systems.

A majority of the established research on creating equitable ride-hailing systems relies heavily on heuristic or linear programming approaches \cite{xu2020trading,suhr2019two,nanda2020balancing,lesmana2019balancing,kumar2023using}, augmented by mechanisms that promote fairness. These traditional algorithms have benefits such as simplicity in structure and reasonable execution times. However, they fall short of guaranteeing non-myopic solutions that can make far-sighted decisions. In addition, existing studies do not utilise longitudinal historical data to identify future patterns for raised requests. These algorithms focus on the immediate future, which lack the predictive capacity for future demand trends and fail to consider historical discrepancies. Raman \textit{et al.} proposed an allocation system based on a Markov Decision Process (MDP) to balance efficiency and fairness, which provides non-myopic allocation plans \cite{raman2021data}. However, the approach to addressing equity issues in the ride-hailing system only considers historical patterns for raised requests, without any consideration for future conditions.

\begin{figure*}
    \centering
    \includegraphics[width=0.7\textwidth]{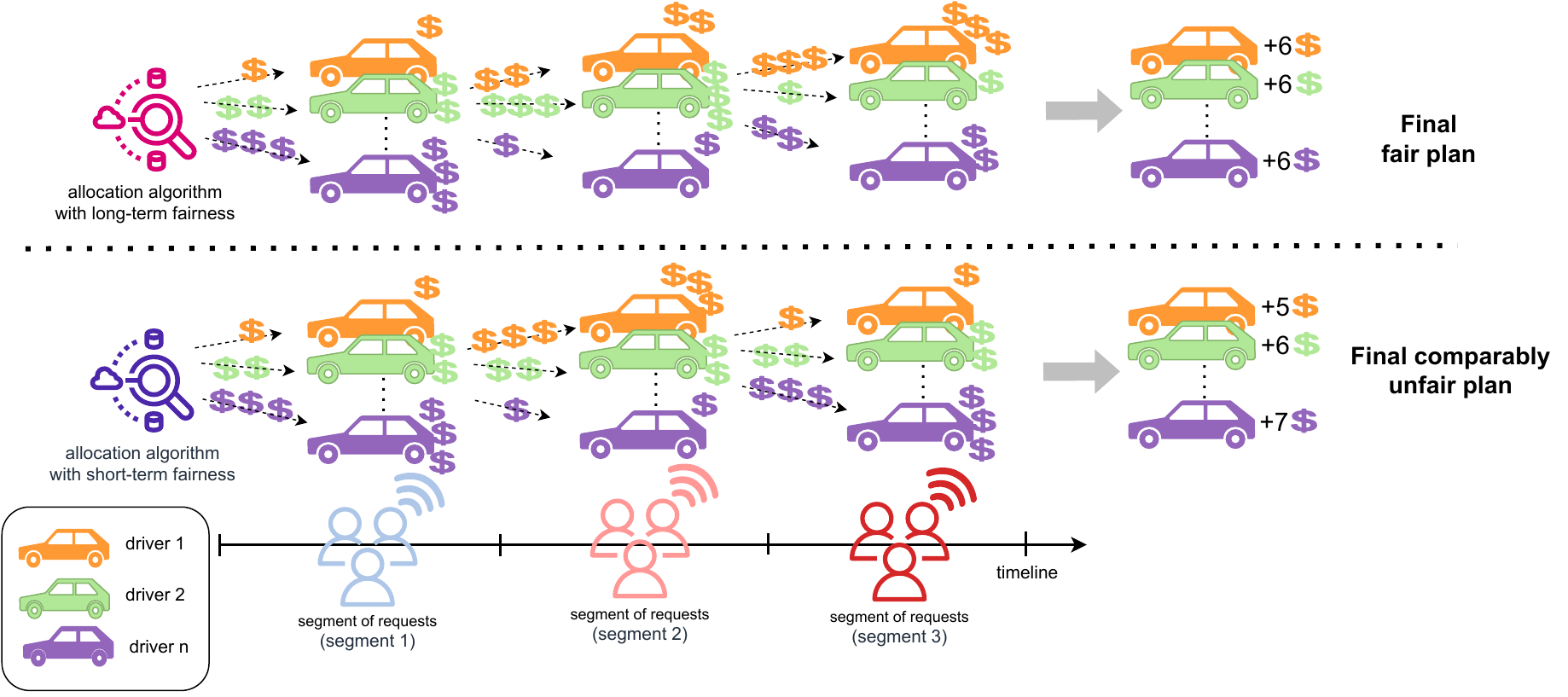}
    \caption{The figure showcases the disparity in fairness concerns between short-term and long-term allocations by displaying two allocation systems generating allocation plans at each timestep. The arrows pointing to vehicles represents allocated requests with different utility to different drivers, and the dollar signs next to drivers indicate the total utility accumulated by the end of each timestep. In this example, the algorithm that prioritizes short-term fairness manages to achieve absolute fairness by the end of the second timestep, with no variance in utility among drivers. However, it becomes unfair by the third timestep. On the other hand, the algorithm that focuses on long-term fairness appears relatively unfair at the end of the second timestep but ultimately achieves absolute fairness by the third timestep.
    }
    \label{fig:scenario}
\end{figure*}

Fig.~\ref{fig:scenario} illustrates a potential scenario where an emphasis on short-term equity results in a long-term inequitable allocation plan. For an allocation system that considers short-term fairness, the system tries to improve fairness by only considering the historical allocations. Thus, at the end of the second timestep, the allocation system achieves fairness among the three drivers, as all the drivers get the same utility, but the final allocation is comparably unfair due to the upcoming requests from the third timestep. As for the allocation system considering long-term fairness, the allocation system is able to consider the future requests patterns and improve fairness across the entire time duration to output a fair allocation plan. Given the existence of multiple possible scenarios where riders raising requests following different distribution, without considering future patterns, long-term equity becomes increasingly unstable, thus underscoring the necessity of a more comprehensive approach. 

Importantly, in practical contexts, drivers in ride-hailing systems are generally more concerned with their weekly earnings, which can be viewed as long-term earnings as opposed to daily earnings \cite{wu2022spatio,huang2022deep}.  The challenge of maintaining long-term fairness can be attributed to three key factors: i) \textbf{Short Sight}: optimisation algorithms without look-ahead time fall short in harnessing long-term historical data and forecasting future patterns. The example shown in Fig.~\ref{fig:scenario} illustrates the challenge. ii) \textbf{Concept Drift}: similar to concept drift in time-series forecasting, existing methods that consider fairness presuppose that future ride requests will adhere to previously observed patterns. However, this assumption is fundamentally flawed as the patterns of ride requests fluctuate continuously due to different factors, including weekday peak hours and public holidays. By its nature, the ride-hailing system necessitates real-time assignments, rendering it unfeasible to accumulate a sufficient volume of ride requests to explore different patterns of raised requests before optimising assignments. iii) \textbf{Disparity between Utility and Fairness Increase across Time}. Multiple fairness definitions have been proposed to evaluate fairness among drivers in ride-hailing systems. An earlier study addressed this by attempting to maximise the minimum utility amongst drivers, treating the problem as an instance of matching issues \cite{lesmana2019balancing}. Nonetheless, the focus of the study on improving fairness is mainly on enhancing the utility for the driver with the lowest utility, thus presenting a rather restricted view of fairness. The study did not consider fairness among all drivers but to improve fairness for the driver with the lowest utility instead. Another approach is to evaluate fairness among drivers based on the variance in travel distance \cite{raman2021data}, which can be considered a comparably more global evaluation of fairness. Research findings indicate a conflict between the optimisation of utility and the promotion of fairness within these systems \cite{xu2020trading,suhr2019two,nanda2020balancing,lesmana2019balancing,kumar2023using,raman2021data,sun2022optimizing}. Therefore, the principal challenge of fairness in ride-hailing lies in reconciling these two objectives. The objective function is normally defined to balance total utility and fairness (e.g., $utility+\lambda fairness$) \cite{raman2021data,lesmana2019balancing,sun2022optimizing}. Whilst we adopt the concept of fairness defined based on variance, we want to highlight the challenge in maintaining this balance over a long time horizon, as the variance inevitably increases, which thus intensifies the preference for fairness. The question of balancing utility and fairness whilst preventing an escalation in the weight of fairness over time remains unresolved.

Therefore, to effectively mitigate these issues, we propose a long-term fairness approach for ride-hailing systems, one that leverages both historical data and forecasts of future scenarios. We propose a Markov Decision Process (MDP)-based method to deliver long-term fairness while balancing it with total utilities for drivers. Our proposed model leverages a multi-objective multi-agent Q-learning (MOMAQL) technique, allowing an allocation plan approach to non-myopic solution that considers various possible data patterns to mitigate the unfairness illustrated in Fig.~\ref{fig:scenario}. In the proposed MDP-based approach, we include a time-series forecasting module which is used to predict future requests, enabling the model to anticipate future patterns. In this case, assuming a driver gets the lowest wage in a short-term time period, the driver can be compensated in the subsequent time periods. 

Our key contributions can be summarised as follows:
\begin{itemize}
\item We formally propose a concept of long-term fairness in ride-hailing systems to promote equal earnings among drivers over a broader time horizon, aligning more closely with their focus on weekly earnings.
\item We present a novel MDP-based model with appropriate objective and scalarisation functions. This model aims to preserve utilities whilst minimising the disparity in earnings among drivers.
\item We introduce a predictive module that forecasts the volume of future ride requests in different locations. This module is incorporated into the MDP-based model to allow a look-ahead time for the MDP-based model and thus help achieve long-term fairness in ride-hailing systems.
\item We validate our proposed model through experiments using a real-world dataset, demonstrating that it outperforms existing methods.
\end{itemize}

\section{Related Work}
The problem of matching ride requests and drivers has been conceptualised as either a bipartite matching problem \cite{zhao2019preference} or a Markov Decision Process (MDP) \cite{raman2021data,sun2022optimizing,10415705}. These studies have demonstrated substantial improvements in efficiency, enabling drivers to service more requests and increase their earnings. However, recent literature has brought equity concerns \cite{rahmattalabi2019exploring}. From the perspective of riders, Brown \textit{et al.} underscored differential treatment of riders by ride-hailing services, leading to an amplified trip cancellation rate for riders with darker skin tones \cite{brown2018ridehail}. Similarly, on the driver side, a subset of ride-hailing drivers struggle to earn a sustainable income due to systemic income disparities \cite{cook2021gender}. To counteract these disparities, some research studies have proposed the use of bipartite matching with a min-max objective function aimed at maximising the minimum utility, thereby promoting more balanced utilities among drivers \cite{lesmana2019balancing}. Subsequent research has demonstrated boundaries on the trade-off between efficiency and a defined concept of fairness \cite{nanda2020balancing,ma2020group}. Upon the aforementioned works, investigations have been carried out focusing on equalising utilities among drivers and riders \cite{suhr2019two,kang2024promoting}. Additionally, Li \textit{et al}. proposed a novel method to augment the fairness of ride-hailing demand functions \cite{yan2020fairness}. Expanding on previous research pertaining to ride-hailing \cite{lesmana2019balancing}, Raman \textit{et al}. formulated a versatile methodology that can be adapted for ride-hailing matching to improve fairness through applying an MDP \cite{raman2021data}. While the work provides a non-myopic solution compared to the previously mentioned studies, the study still focuses on promoting short-term fairness like other studies without an estimation of patterns of future requests.

\section{Preliminaries}
Studies revealed that there is a conflict between fairness and utility, and optimising fairness without considering utility will cause all the drivers not to get any assignment (no earnings for everyone is absolutely fair) \cite{suhr2019two,nanda2020balancing,shi2021learning}. Thus, to consider fairness in ride-hailing, the target is to balance between utility and fairness. 

\subsection{Problem Formulation}

Assume there are $n$ drivers $v\in V$ who aim to pick up different orders sent by riders, and all drivers and riders reside on a 
directed graph. The set of locations, $L$, are represented as nodes in this directed graph which can be locations for the $n$ drivers or either pickup or drop-off locations for rider-raised requests. The set of directed edges, $E$, where $e_{i,j}\in E$, represents the travel distance between locations $l_{i}, l_j\in L$. In the real-world scenarios, it is normal that the travel distance from a location (e.g. $l_i$) to another location (e.g., $l_j$) is different from travelling in the opposite direction. Thus, we define the real-world map as a directed graph. We define the utility of a trip between the location $l_{i}$ and $l_{j}$ based on $e_{i,j}$ according to existing studies \cite{raman2021data,tong2020spatial}. 

The rider requests, driver states and time length are formulated as follow:

\begin{itemize}
    \item A rider request $r\in R$ and $r = (t_r,s_r,d_r)$ where the request $r$ is raised at time $t_r$ from locations $s_r$ to $d_r$; $s_r, d_r\in L$. 
    \item The state of each of the $n$ drivers at time $t$ can be represented by the tuple $v^t\in V^t$ and $v^t = (c_{v}, m^{t}_{v}, g^{t}_{v}, o^{t}_{v})$, where $c_{v}$ represents the capacity of the vehicle that driver $v$ is driving, $m^{t}_{v}$ is the number of riders in the vehicle at time $t$, $g^{t}_{v}$ is the geographical location of $v$ at time $t$ ($g^t_{v}\in L$) and $g^t_{v}$ is dynamically changing based on $t$ with $g^t_{v}=-1$ if the driver is travelling on an edge between two different nodes, $o^{t}_{v}(R_v)$ is the total utilities gained by $v$ from the requests that are assigned to $v$, represented by $R_v$, from starting time $t_0$ to $t$.  
    \item The time length $T=\{\{t_{0-\delta } ,\ t_{0-\delta +1} ,\ ...,\ t_{0-1}\} ,\ t_{0} ,\ \{t_{1} ,\ t_{2} ,\ ...,\ t_{n}\}\}$ of the problem is formulated in three parts: the time length $\{t_{0-\delta } ,\ t_{0-\delta +1} ,\ ...,\ t_{0-1}\}$ when the historical orders are already completed or being processed, the time step $t_0$ when the current orders have been raised but not been assigned to the drivers, and the time length $\ \{t_{1} ,\ t_{2} ,\ ...,\ t_{n}\}$ when the future orders will need to be assigned to the drivers. 
\end{itemize}

It is important to note the utility of each request $r$ is calculated by $Geo(d_r, s_r) - Geo(s_r, g^t_v)$ where $Geo$ calculates the shortest geographical distance between two points. The utility reflects the balance between profit, defined as the distance from the start to end location for a request ($s_r$ to $d_r$) and cost, represented by the distance from the current location of the driver ($g^t_v$) to the start location of the request ($s_r$). The utility calculation is performed at the moment a request is assigned. We assume drivers opt for the shortest possible route, meaning the utility is consistent for requests with identical start and end points. Upon assignment of a request $r$ to a driver $v$, its utility is added to the cumulative utility of the driver $o^{t}_{v}$. 

\subsubsection{Requests Prediction}
Assume a set of requests raised by different riders across the time length $T$ is represented by $R_{T} =\{\{R_{t_{0-\delta }} ,\ R_{t_{0-\delta +1}} ,\ ...,\ R_{t_{0-1}}\} ,\\ R_{t_{0}} ,\ \{R_{t_{1}} ,\ R_{t_{2}} ,...,\ R_{t_{n}}\}\}$ where $\{R_{t_{1}} ,\ R_{t_{2}} ,...,\ R_{t_{n}}\}$ represents the future orders that will be raised by different riders. $R_{t_{n}} =\{(s_{r}^{t_{n}} ,d_{r}^{t_{n}} )|s\in L,\ d\in L\}$ represents orders that will be raised at node $s\in L$ and end at node $d\in L$ at time $t_n$. We aim to forecast the number of trips raised and ends at each node in the graph.

\subsection{Efficiency}

This study defines overall efficiency as the total utility acquired across all the drivers over a given timeframe. Given a set of requests $r\in R$, efficiency aims to find an assignment $M$ that assigns each request to exactly one driver $v\in V$ to maximise total utility shown as: 

\begin{equation}
   \begin{gathered}
        \pi(M)=\sum_{v \in V} o^{t_n}_{v}(M(v)) \\
        t_n=\max(T)
    \end{gathered}
\end{equation}
where $M(v)$ represents the set of requests that are assigned to the driver $v$ according to the assignment $M$.

\subsection{Long-term Fairness}

We define long-term fairness as the accumulated fairness that consider fairness calculated from the historical, current, and future allocation plan which will be the output from the proposed method using testing data. In this study, we use 1 week as the time horizon with the first three-days records treated as historical data, the fourth day treated as the current data, and the last three days treated as the future data which will be considered as testing data for the proposed method. The fairness is calculated as the accumulated fairness through the whole week as drivers care more about weekly-based earnings instead of daily earnings in the real-world. The fairness is defined as the variance of utilities among the $n$ drivers according to existing studies \cite{raman2021data}, where utilities for each driver is defined as the total utilities through the whole week. Given a set of requests $r\in R$, long-term fairness aims to find an assignment $M$ that assign each request to exactly one driver $v\in V$ to minimise long-term fairness shown as: 

\begin{equation}
     \begin{gathered}
        F(M)=\operatorname{Var}\left(o^{t_n}_v(M(v))\right) \\
        t_n=\max(T)
    \end{gathered}
    \label{fairness}
\end{equation}
where $\operatorname{Var}$ represents variance to calculate the variance of total utilities among different drivers.

\subsection{Balance of Long-term Fairness and Total Utility}
To maintain a balance between long-term fairness and total utility, we formulate the problem as an operation research question with the formula shown as:

\begin{align}
    \label{optimisation}
    \begin{split}
       max_M\ \uppi(M) - \lambda F(M)
    \end{split}
        \shortintertext{subject to}
        &\sum_{v\in V, r\in R} I_{rv} \leq 1
\end{align} 
where $\lambda$ is the weight for fairness, and $I_{rv}$ is defined as an indicator function:

\begin{equation}
    I_{rv} \ =\ \begin{cases}
    1 & if\ request\ r\ is\ assigned\ to\ vehicle\ v\\
    0 & else
    \end{cases}
\end{equation}

\section{Approach: Optimising Efficiency and Long-term Fairness for Ride-Hailing}

In this section, we introduce a novel solution to both optimise the efficiency and longer-term fairness for ride-hailing applications. The proposed model adopts a multi-objective multi-agent Reinforcement Learning (MOMARL) algorithm to develop the allocation system, driven by three core considerations:

\begin{itemize}
    \item \textbf{Real-World Dynamics and Initial Conditions}: The varying initial locations of drivers in the real-world significantly influence their ability to serve ride requests, affecting their behaviour and accumulated utility. MOMARL allows different drivers learn different agent behaviours based on their starting locations, thereby optimising total income while promoting fairness.
    \item \textbf{Fairness and Equity Considerations}: To address fairness, defined as the equitable comparison of utility gained by drivers, our system employs a central controller to acknowledge status of all agents. This ensures that allocation decisions consider the collective situation of all drivers, promoting a fair distribution of utility and opportunities across the network. The central controller allows the proposed method dynamically adjusts to real-time conditions and redistributes resources to maintain fairness and efficiency.
\end{itemize}

We designed the scalarisation function in MOMARL to balance utility and fairness and to approach Pareto Optimal. Additionally, a time-series prediction module is incorporated to provide future available actions in MOMARL to allow the proposed method to consider future requests' pattern. The prediction module is implemented to adapt the dynamics of the requests raised by riders based on time in an online manner.

\subsection{Overview}

\begin{figure}
    \centering
    \includegraphics[width=0.6\textwidth]{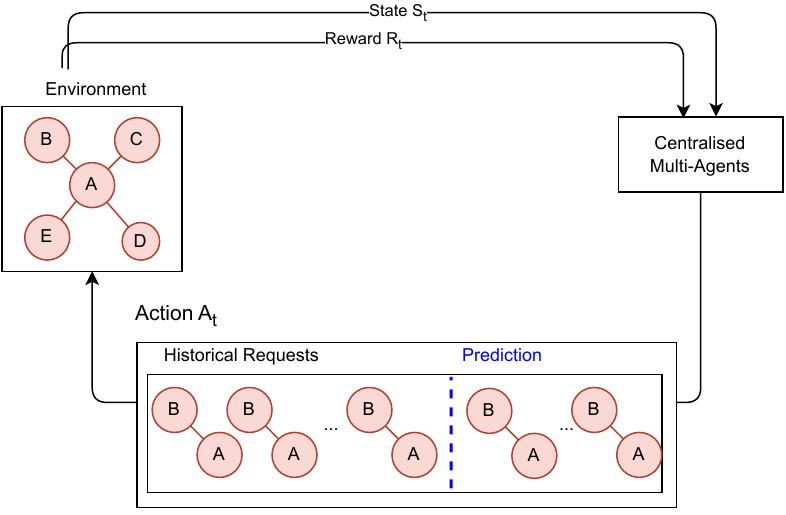}
    \caption{Long-term Fairness for ride-hailing system. With time-series prediction, the predicted requests is part of the action space of the MDP-based model to allow the outputed allocation plan be based on the pattern of future requests.}
    \label{fig:my_label}
\end{figure}

The proposed model comprises a time-series prediction module, multi-objective Reinforcement Learning, and a scalarisation function. 

\begin{itemize}
    \item Time-series forecasting (Sec.~4.2). To encourage longer-term fairness, we utilise time-series forecasting in the proposed model to predict future requests as part of the input for MOMAQL. The time-series forecasting module exploits historical requests in different locations as input and the output is the number of requests that will be raised in the future from different locations. 
    
    \item Multi-objective multi-agent Reinforcement Learning (Sec.~4.3). In this study, we exploit multi-objective multi-agent Q Learning (MOMAQL) to construct the fundamental part of the proposed model as Reinforcement Learning has been proven to be an efficient method to construct ride-hailing systems \cite{de2020efficient}. For each time step, the centralised controller assigns each request to an agent (a driver $v\in V$) if $g^t_v\neq -1$. Each objective function of MOMAQL focuses on maximising the utilities for each driver, where the utility of each request is calculated based on the geographical shortest distance. In this way, MOMAQL here output an allocation plan target on maximising the total utilities among different drivers. 
    
    \item Scalarisation function (Sec.~4.4). In order to transform the multi-objective problem into a standard single-objective problem, we then propose a scalarisation function. The scalarisation function not only aims to transfer the problem, but by maximising the value of the scalarisation function, it also seeks for a balance between efficiency and fairness to approach Pareto Optimal.  
\end{itemize}

The proposed model operates in four stages in each batch: predicting, evaluating, assigning and learning. We first predict the number of future requests in different locations. Then, each batch starts with an evaluation: when a request is raised by a rider, MOMAQL first finds the shortest path from the start location (the location where the request is raised) to the end location (the required destination), where the travel distance represents the utility of the request. For the assigning stage, MOMAQL assigns the request to a driver based on the value calculated by the scalarisation function, considering both efficiency and fairness. Lastly. the proposed MOMAQL based model learns from the matching results and utilises the scalarisation function to optimise both the efficiency and fairness and approach Pareto Optimal.

\subsection{Time-series Forecasting}

The request prediction module is defined based on time-series prediction for which we utilise Multi-Layer Perceptron in this study (MLP). MLP consists of an input layer of source nodes, one or more hidden layers, and an output layer. As an existing study stated that a single hidden layer is sufficient to approximate different continuous functions \cite{greenwood1996guide}, we use a three-layer MLP in our proposed method. We first utilise the pairs of locations (start  and destination locations from different requests) as features, then multiple measurements at time $t, (t-1), ..., (t-n)$ are used to predict the requests that will happen in the future (the 7 days that we use to test the model), where each time step is set as 1 hour. The structure of the request prediction module has $number$ of neurons in the hidden layer. By using the chosen dataset, we use the previous 1 month of data for training and output the number of requests that will happen based on each pair of locations in the next 7 days.   

\subsection{Multi-Objective Multi-Agent Q Learning}

\begin{figure}
    \centering
    \includegraphics[width=0.7\textwidth]{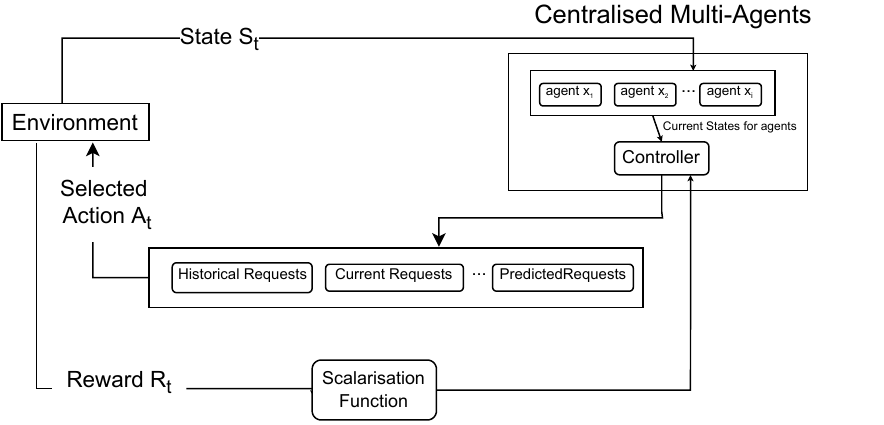}
    \caption{Multi-objective multi-agent Q Learning. By customising the action space and scalarization function, we aim to encourage the balance between utility and fairness by utilising multi-objective multi-agent Q learning. In action space, it includes historical, current and predicted future requests to allow the proposed model trained based on the pattern of future requests. For scalarisation function, it is designed aiming to balance utility and fairness and maximise the objective.}
    \label{fig:my_label}
\end{figure}

To convert the ride-hailing assignment problem to a Markov Decision Process (MDP),  we define its foundational elements as follows:

\begin{itemize}
    \item \textbf{State }: The states are derived from the various locations where drivers are initially positioned and where ride requests originate and conclude. These locations constitute a finite set of states, encapsulating both driver positions and request locations at any given time point $t$.
    \item \textbf{Action}: Within the MDP framework, actions at each time point $t$ where $t\in T=\{\{t_{0-\delta } ,\ t_{0-\delta +1} ,\ ...,\ t_{0-1}\} ,\ t_{0} ,\ \{t_{1} ,\ t_{2} ,\ ...,\ t_{n}\}\}$, represent the strategic assignment of incoming ride requests to available drivers. The driver is requested to drive from the current location to the start location of the request to pick up the rider and drive to the destination. The proposed method allows an agent (driver) to accept multiple requests concurrently. The time series forecasting module is incorporated into the MDP framework to predict available actions at $t\in T$, which enables the anticipation of future requests and their strategic incorporation into current decision-making. This forward-looking capability ensures that actions not only respond to immediate demands but also adapt to predicted future conditions. Additionally, the model supports the option of no action at $t$, allowing periods where a driver may not receive any assignments, which provides a mechanism to prevent overburdening drivers and ensuring a fair distribution of work.
    \item \textbf{Reward}: Once each agent (driver) takes action, the agent receives an instant reward shown as: 
    \begin{equation}
    r_{s, A}^{t} (v)\ =\ \sum _{a \in A_{v}^{t}} Geo(d^t_{a}, s^t_a) - Geo(s^t_a, g^t_v)
    \end{equation}
    where $A^t_v$ represents the set of actions taken by the agent $v$ at time $t$, $s^t_a$ represents the starting location of the assigned request, $d^t_a$ represents the ending location of the assigned request, and $g^t_v$ represents the location of $v$ which means the state of the agent at time $t$. 
\end{itemize}

The proposed model exploits centralised MOMAQL. For each time step, the centralised controller queries the raised requests and assigns requests to different agents that is currently located on a node. To simplify the problem, we set the drivers always to accept requests without cancellation. 

\subsection{Scalarisation Function}
One of the approaches for multi-objective problems relies on single-policy algorithms \cite{miettinen2002scalarizing} to learn Pareto Optimal solutions. Single-policy multi-objective Reinforcement Learning algorithms exploit scalarisation functions over the vector-based reward functions, thereby reducing the multi-objective environment's dimensionality to a single, scalar dimension. Maintaining a balance between optimising utility and improving fairness over a long time horizon is challenging as the two objectives increase at different speeds. Thus we define the scalarisation function with a weight for fairness to adjust its range. The scalarisation function is designed based on Eq.~\ref{optimisation} as:


\begin{equation}
     \begin{array}{ l }
        \text{SR}(M)\ =\ \sum _{v\in V} r_{s,A} (v)\ -\ \lambda \omega \operatorname{Var}(r_{s,A} (v)) \ \\
        r_{s,A} (v)\ =\ \sum _{t\in T} r_{s,A}^{t} (v)\ ,
    \end{array}
    \label{scalarisation}
\end{equation}
where $0\leq \lambda \leq 1$ represents the weight for fairness among different drivers and $0<\omega \leq 1$ represents the scale to adjust fairness into the same range as utility. By adjusting $\omega$, the weight assigned to fairness is adjusted to avoid fairness getting a larger weight due to the unavoidable increase of variance while the time horizon gradually increases. 

\section{Experiments}

\begin{table*}[t]
\caption{Long-term fairness performance compared with baselines on the dataset.}
\label{results_table}
\resizebox{\textwidth}{!}{
\begin{tabular}{lllllll}
\hline
\textbf{Methods}         & \textbf{Total Utility} & \textbf{Fairness} & \textbf{Normalised Fairness} & \textbf{Min} & \textbf{Mean} & \textbf{Max}  \\ \hline
Greedy          & -1514736.24   & 1696.95 & -0.0005 & -75803.61 & -75736.81 & -75653.78   \\ \hline
REASSIGN        & 76536.23      & 493637.57 & 0.18 & 2218.72 & 3826.81 & 4760.17 \\ \hline

LAF & 80606.49 & 107789.96 & 0.0814 & 3001.74 & 4030.3245 & 4491.26 \\ \hline
Balance Ride-Pooling & 85923.68      & 100254.73 & 0.074 & 3451.47 & 4296.18 & 4674.44 \\ \hline
Proposed Method & 95823.79      & 85194.48 & 0.061 & 4565.56 & 4791.19 & 5931.93  \\ \hline
\end{tabular}
}
\end{table*}

\begin{figure}[!ht]
    \centering
    \includegraphics[width=0.65\linewidth]{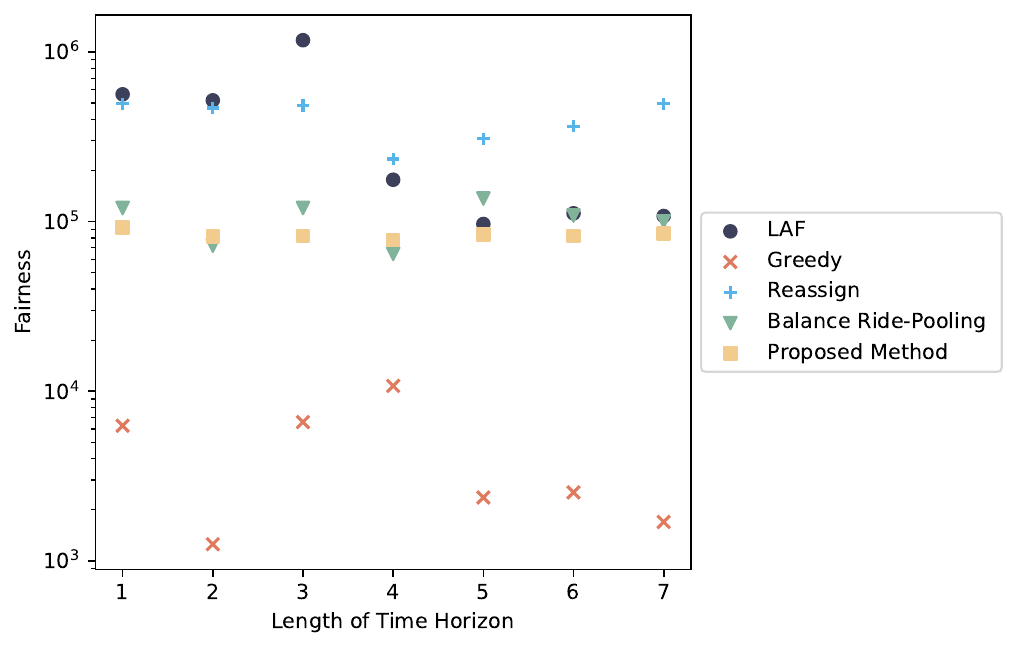}
    \caption{Performance of baselines and proposed model in terms of fairness based on gradually increased time horizon}
    \label{fig:long_term_stablness_fairness}
\end{figure}

\subsection{Datasets}

We exploit New York City Taxi dataset\footnote{https://www.nyc.gov/site/tlc/about/tlc-trip-record-data.page}, a publicly available taxi trip dataset collected in New York City, which contains essential information on all requests executed by active taxis including the day and time when a request is raised, the location where the request is raised, and the dropoff location it required. Each request's pickup and drop-off locations are recorded in longitude and latitude coordinates. We extract all requests starting and ending within Manhattan, happening on the dates ranging from 01/03/2016 to 01/04/2016. For simplicity, the shortest travel time from a certain pickup to a certain drop-off location is re-calculated as the mean travel time across the time period, and multiple locations are merged together as a node in constructed graph. This study assumes that the driver will always choose the shortest path to complete a request.  

\subsection{Experimental Details}

\subsubsection{Baselines}

We select three existing fair ride-hailing methods and Greedy with the objective to balance efficiency and fairness as baselines:

\begin{itemize}
    \item Greedy. We implement Greedy with the objective of balancing efficiency and fairness according to Eq.~\ref{optimisation}.
    \item REASSIGN. REASSIGN exploits traditional optimisation with the objective of balancing efficiency and fairness. To compare with our proposed method, the fairness definition is modified in REASSIGN according to Eq.~\ref{fairness}. In the study, Lesmana \textit{et al.} state that their proposed method can be applied with various fairness definitions \cite{lesmana2019balancing}. 
    \item LAF. The study conducted by Shi \textit{et al.} exploits a Markov Decision Process as a re-weighting module to refine the weight for each edge to promote fairness. LAF then utilise Hungarian algorithm to optimise total utility and output the final allocation plan. LAF is used as one of the baseline with the fairness definition modified according to Eq.~\ref{fairness} \cite{shi2021learning}.  
    \item Balance Ride-Pooling \footnote{The study has not provided a formal name}. The study conducted by Raman \textit{et al.} exploits a Markov Decision process to optimise the number of rider requests serviced while maintaining fair earning among drivers. We use the method proposed by Raman \textit{et al.} as one of our baselines \cite{raman2021data}. The fairness definition they used in their study is similar to the fairness definition in our study. Hence, the implementation remains unchanged. 
\end{itemize}

\subsubsection{Experimental Settings}
We selected data before 26/03/2016 as the training data and predicted the requests from 26/03/2016 to 01/04/2016. To reduce the training time, we extracted peak 2-hour data ranging from 19/03/2016 to 01/04/2016 and for the request prediction output. We then use the first seven days data and the extracted output from time-series forecasting to train our proposed model and test the remaining data. During the training process, we used a stratified sampling method with a sampling rate of 0.05 for the training data. We set $\lambda=1$ (a parameter shown in Eq.~\ref{scalarisation}) to indicate no preference on utility or fairness, $\omega=0.6$ (a parameter shown in Eq.~\ref{scalarisation}) to scale utility and fairness into the same range, $\gamma=0.9$ as the value for discount factor for MOMAQL. All experiments are trained and tested on a Linux system (CPU: Intel(R) Xeon(R) Gold 6240 CPU @2.60GHz, GPU: NVIDIA GeForce RTX 8000).

\subsection{Results and Analysis}
As the range of variance varies significantly based on the attained total utility, we further utilise normalised fairness as another measurement to show experimental results. In this study, we define normalised fairness as normalised standard deviation among the utilities for different drivers shown as: 

\begin{equation}
    \hat{F}(M) \ =\ \frac{\sigma (U )}{\overline{U}},
\end{equation}
where $U$ represents vector records accumulated utility by each driver and $\overline{U}$ represents the mean utility across all drivers.

Table~\ref{results_table} and Fig.~\ref{fig:long_term_stablness_fairness} summarise the results of the proposed method compared to the baselines on the real-world dataset. We further tested the performance of each method by gradually increasing the prediction horizon to test the stability of fairness in terms of the time horizon. All the experiments are conducted under the same experimental settings.

\subsubsection{Long-term Fairness Performance Comparison}
Under this setting, each method attains an optimised allocation by using the whole seven days testing data. For Table~\ref{results_table}, we can observe that: (1) The two objectives, fairness and efficiency, are contradicted to a certain level. For Greedy, with the optimised allocation focusing more on fairness, the total utility can even reach negative, which cannot be a solution in the real-world. In order to obtain a fair result, Greedy tends to sacrifice the utility of all the other drivers to achieve a fair result based on the driver with lowest utility instead of increasing the utility for the driver. Essentially, all the drivers are allocated to the requests with the lowest utility which leads to the final result with the negative value for efficiency. (2) Comparing REASSIGN (based on traditional optimisation) with the Proposed Method (based on Reinforcement Learning), we can see that the proposed method perform better, which further supports the statement made by Shah \textit{et al.} \cite{shah2020neural} that traditional optimisation makes comparably more myopic decisions compared to Reinforcement Learning. (3) Balance Ride-Pooling proposed by \cite{raman2021data} is also based on Reinforcement Learning, Compared with our proposed model, the method does not consider the patterns of requests in the future, which indicates the dependency of future patterns can improve the total utility for the drivers. (4) The proposed method achieves more balanced results in terms of efficiency and fairness compared to Greedy and outperforms other baselines. 

\subsubsection{Stability of Long-term Fairness}
We aim to compare existing methods with our proposed model in terms of fairness with various time horizons. Fig.~\ref{fig:long_term_stablness_fairness} shows that: (1) With predicted future patterns, fairness is lower when the length of the time horizon is 1 and gradually improves with increases in the time horizon and approaches a stable value after the length of the time horizon is equal to 4. This is our target as long-term fairness focuses on the stability and value of fairness when the time horizon is longer, as drivers will care more about the equity of profits in the long term compared to the short term. (2) Comparing REASSIGN with Balance Ride-Pooling and the Proposed Method, it shows that the output from Reinforcement Learning based methods are comparably more stable when the length of the time horizon increases. (3) Compared with Balance Ride-Pooling, our proposed method includes the requests prediction, which allows the proposed model output allocation plan with dependency on future situations to achieve a more stable result for fairness.

\subsection{Ablation Study}
\begin{figure}[!ht]
    \centering
    \includegraphics[width=0.7\linewidth]{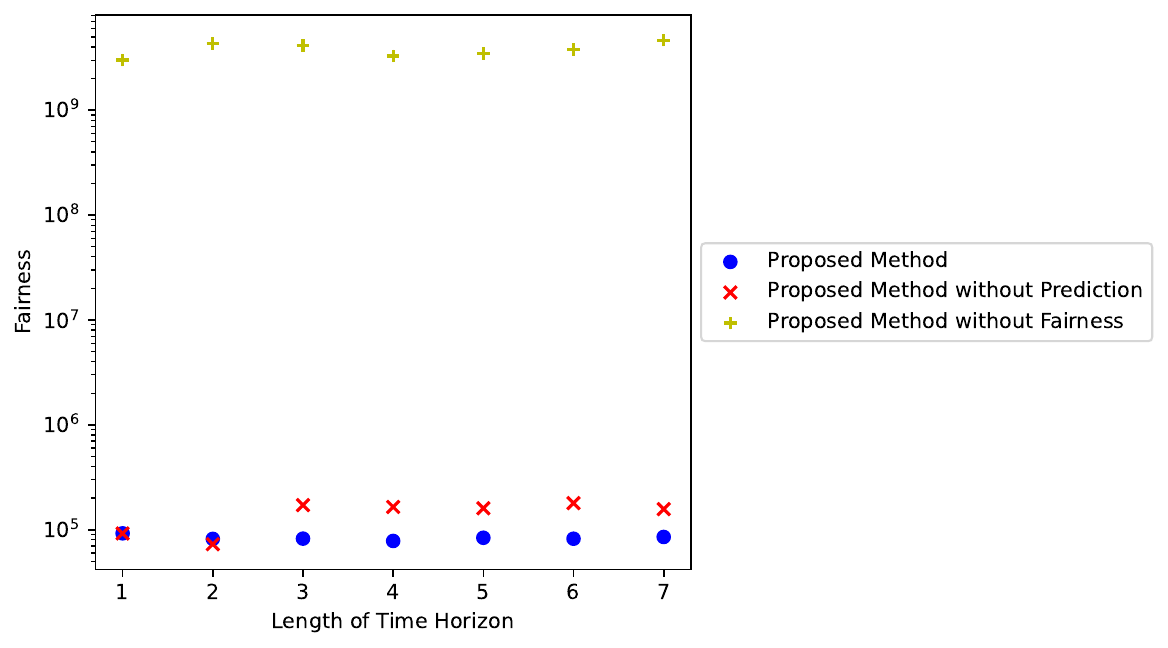}
    \caption{Ablation study. Performance of the proposed model without different modules in terms of fairness and gradually increased time horizon, where the time horizon is increased by a number of days. For fairness, the larger value indicates the model is unfairer. }
    \label{fig:abalation_long_term_stablness_fairness}
\end{figure}

Table~\ref{abaltion_results_table} and Fig.~\ref{fig:abalation_long_term_stablness_fairness} show the ablation study to answer two following questions:

\textbf{How well does the request prediction module balance total utility and fairness?} 
\begin{table}[t]
\caption{Ablation study. Long-term fairness performance compared with the proposed method excluding different modules on the selected dataset.}
\label{abaltion_results_table}
\centering
\begin{tabular}{lll}
\hline
Methods         & Total Utility & Fairness  \\ \hline
Our Method & 95823.79      & 85193.62  \\ \hline
Our Method w/o Prediction        & 56873.21      & 153697.27 \\ \hline
Our Method w/o Fairness         & 2194901.19   & 2677473902.27   \\ \hline
\end{tabular}
\end{table}
In this study, Table~\ref{abaltion_results_table} shows the performance comparison of the proposed method, proposed method without prediction, and proposed method without fairness. The proposed method without fairness does not consider fair earning among different drivers but leads to the highest total utility. By considering fairness, comparing the proposed method with the proposed method without fairness, it shows the request prediction module not only encourages fairer earning among different drivers but also increases the total utility. The Mean Squared Error of the request prediction module is 94.69, and the reason for the improvement on total utility and fairness is that the proposed method utilises the predicted future requests in the training process, which further updates the Q-Table based on future patterns of the requests.  

\textbf{How well does the request prediction module work on the stability of fairness with gradually increasing time horizons?} Fig.~\ref{fig:abalation_long_term_stablness_fairness} shows the performance of the stability of fairness with gradually increasing time horizons on the proposed method, proposed method without prediction, and proposed method without fairness. For the proposed method without fairness, as the method does not consider fairness in the allocation, the fairness is comparably unstable with high unfairness. The earnings instability among different drivers can cause issues in the real world. Comparing the proposed method and the proposed method without a prediction module, the proposed method without the prediction module achieves fairer results when the length of the time horizon is 1 and 2. As the length of the time horizon increases to 3 or more, the fairness of the proposed method is better than the proposed method without prediction and more stable. It shows the prediction module helps to achieve fairer and more stable results in the long-term with sacrifice on the short-term, and we argue that long-term fair earning is what the drivers desire in reality.

\section{Conclusion}
In this paper, we formally proposed long-term fairness, which focuses on achieving stable and comparably higher fairness over comparably longer time horizons. We argue that taxi drivers will care more about long-term earnings. To achieve the target, we introduce a request prediction module before allocation to allow look-ahead windows for the proposed allocation system and eliminate the assumption that requests always follow the same pattern. We exploit the output as part of the action space for the allocation model, which we designed using Multi-objective Multi-agent Q Learning. The experiments on real-world data demonstrated the effectiveness of our proposed method for maintaining overall fairness in the comparably longer time horizon and enhancing the stability of fairness when the time horizon gradually increases.


\begin{credits}
\subsubsection{\ackname} We acknowledge the support of the Australian Research Council (ARC) Centre of Excellence for Automated Decision-Making and Society (ADM+S) (CE200100005). This research was partially supported by NVIDIA Academic Hardware Grant Program. This research was partially supported by Google Travel Grant. This research was undertaken with the assistance of computing resources from RACE (RMIT AWS Cloud Supercomputing).

\end{credits}
\bibliographystyle{splncs04}
\bibliography{ecmlpkdd}

\end{document}